# Embedding Textual Information in Images Using Quinary Pixel Combinations

A V Uday Kiran Kandala, Queen Mary University of London
a.kandala@se25.qmul.ac.uk

**ABSTRACT:** This paper presents a novel technique for embedding textual data into images using quinary combinations of pixel intensities in RGB space. Existing methods predominantly rely on least and most significant bit (LSB & MSB) manipulation, Pixel Value Differencing (PVD), spatial perturbations in RGB channels, transform domain based methods, Quantization methods, Edge and Region based methods and more recently through deep learning methods and generative AI techniques for hiding textual information in spatial domain of images. Most of them are dependent on pixel intensity flipping over multiple pixels, such as LSB and combination of LSB based methodologies, and on transform coefficients, often resulting in the form of noise. Encoding and Decoding are deterministic in most of the existing approaches and are computationally heavy in case of higher models such as deep learning and gen AI approaches. The proposed method works on quinary pixel intensity combinations in RGB space, where five controlled different pixel intensity variations in each of the R, G, and B channels formulate up to one hundred and twenty five distinct pixel intensity combinations. These combinations are mapped to textual symbols, enabling the representation of uppercase and lowercase alphabetic characters, numeric digits, whitespace, and commonly used special characters.Various textual information is successfully embedded into different sizes of images and retrieved without any deviations. Different metrics such as MSE, MAE, SNR, PSNR, SSIM, Histogram Comparison and Heatmap analysis, were evaluated for both original and encoded images resulting in no significant distortion in the images. Furthermore, the method achieves improved embedding efficiency by encoding a complete textual symbol within a single RGB pixel, in contrast to LSB and MSB based approaches that typically require multiple pixels or multi-step processes, as well as transform and learning based methods that incur higher computational overhead.

## 1. INTRODUCTION:

The widespread use of digital images as a medium for information exchange has motivated extensive research into techniques for embedding data within images. Image steganography aims to hide textual or binary information inside digital images while preserving visual quality and avoiding detectable artifacts. In contrast to cryptographic approaches, which protect the content of a message but explicitly indicate that communication is taking place, steganographic techniques focus on concealing the very existence of the embedded data. As a result, image steganography has become relevant in scenarios such as secure communication, data authentication, copyright protection, and privacy-aware information sharing across open networks. Among the various forms of data embedding, hiding textual information within images remains a fundamental and widely studied problem due to the structured and symbolic nature of text data. Text-based steganography requires reliable symbol representation, accurate decoding, and minimal visual distortion, especially when embedding longer messages. Achieving this balance between embedding capacity, imperceptibility, and decoding correctness continues to be a key challenge in image-based text hiding methods. Existing approaches for hiding textual data in images predominantly operate in the spatial or transform domains [17]. Spatial-domain techniques commonly embed text by modifying least or most significant bits of pixel intensities, or by exploiting pixel value differences between neighboring pixels [22]. Transform-domain methods embed textual information within frequency coefficients obtained using transforms such as DCT or DWT, while adaptive and edge-based techniques attempt to place data in visually complex regions to reduce perceptibility [23]. More recently, learning-based and generative approaches have also been explored for text hiding, though these methods typically introduce higher computational complexity and reduced interpretability [13]. A substantial portion of image steganography research has been centered on bit-level spatial-domain embedding, particularly least significant bit (LSB) substitution and its extensions. Works by Kordov and Zhelezov [1], Sakshi et al. [2], Shukla et al.[3], Banerjee et al.[4], Halboos and Albakry [5], Nazmun Nahar et al. [6], May Alanzy et al. [7] and Iman et al. [21] embed textual information by modifying one or more LSBs of RGB pixels, often combined with encryption, hashing, chaotic pixel traversal, or adaptive region selection to improve security and imperceptibility. While these enhancements reduce detectability, such approaches remain fundamentally bit-centric, typically requiring multiple pixels or multi-bit aggregation to represent a single textual character, thereby increasing payload dispersion and susceptibility to pixel level statistical and histogram based steganalysis, especially under higher embedding rates or image post-processing.

Several studies move beyond plain LSB substitution by introducing block based or indirect symbol encoding strategies. Jassim's Five Modulus Method (FMM) [8] encodes text by constraining pixel values within fixed windows and using residue positioning to represent characters, achieving good visual quality by exploiting human visual tolerance to small intensity changes. However, this approach depends on window level processing and indirect symbol localization, introducing multi pixel dependency and limiting symbol level efficiency.

Similar limitations arise in hybrid schemes that combine lightweight transforms or mathematical mappings with spatial embedding, where characters are still reconstructed indirectly through pixel groups rather than being encoded explicitly within individual pixel intensity states. More recent work explores transform-domain, learning-based, and generative steganography frameworks. Srinivasu and Veeramani [9], Lan et al. [10], Ding et al. [11], Miran Hikmat Mohammed [20] and Kumar et al. [12] employ wavelet or DCT transforms, convolutional neural networks, adversarial perturbations, or invertible neural networks to distribute hidden information across spatial or frequency representations, achieving high imperceptibility and robustness against compression and noise. Jingyun et al. [13] further extend this paradigm through generative diffusion based image synthesis, embedding messages directly during image generation. Although effective, these methods rely on trained neural models, complex optimization procedures, or generative inversion, leading to high computational cost, non-deterministic decoding, limited reproducibility, and reduced practicality for lightweight or real time steganographic communication. In [14] alphabet level symbol combinations have also been explored in raster domain text steganography to encode textual information through controlled perturbations.

Comprehensive surveys by Mandal et al. [15]., Kumar et al. [16]., Bhatt and Savant [17], Ragab et al. [18], Manndy and Sebastian [24] and Alenizi et al. [19] systematically evaluate spatial-domain, transform-domain, cryptographic, adaptive, and deep learning based steganography techniques, consistently highlighting the trade-offs between capacity, imperceptibility, robustness, and complexity. Across these surveys, a recurring observation is that most existing methods depend on multi-pixel bit manipulation, transform coefficients, or encryption heavy pipelines, and that symbol-level embedding efficiency remains underexplored in the spatial domain. Complementing this perspective, Knöchel and Karius [20] demonstrate through extensive analysis of text steganography and Stegomalware that real world deployments overwhelmingly favor simple, deterministic, low complexity embedding strategies over complex learning based schemes. Together, these findings motivate the need for direct, deterministic pixel level symbol encoding mechanisms that can increase embedding efficiency while preserving visual fidelity and computational simplicity. To overcome the limitations of bit-level and multi-pixel embedding schemes, this work introduces a quinary pixel combination based image steganography approach that encodes textual symbols directly within controlled RGB intensity variations of a single pixel. Each character is mapped to a valid quinary offset combination across RGB color channels, ensuring deterministic encoding and decoding through pixel wise comparison with the original cover image. By avoiding cryptographic preprocessing, transform-domain operations, and learned models, the proposed method achieves higher symbol-level embedding efficiency, low visual distortion, and lightweight computation, making it suitable for practical and resource efficient hidden communication.

## 2. METHODOLOGY:

The proposed methodology introduces a deterministic spatial domain text embedding scheme based on quinary pixel intensity combination in RGB space, designed to encode complete textual symbols within a single color pixel. Instead of relying on bit-level substitution or multi-pixel aggregation, the method represents characters through controlled, bounded intensity variations applied independently to the R, G, and B channels. By restricting each channel to five permissible perturbation levels, namely -2, -1, 0, +1, and +2, the combined RGB space yields 125 unique and distinguishable pixel-state combinations, enabling direct symbol-to-pixel mapping.

Encoding is performed by selectively perturbing pixel intensities relative to the cover image, while decoding is achieved through difference based analysis using the original image as reference, ensuring deterministic and lossless symbol recovery. The methodology directly operates within constrained intensity bounds to maintain visual imperceptibility and prevent overflow or underflow in pixel values. By treating pixel intensity differences as symbolic carriers rather than binary containers, the approach shifts steganography from a bit-centric paradigm to a symbol-level representation in continuous pixel space, offering higher embedding efficiency with minimal structural distortion. The perturbation mechanism and the resulting quinary combination space are illustrated in the following figure 1.

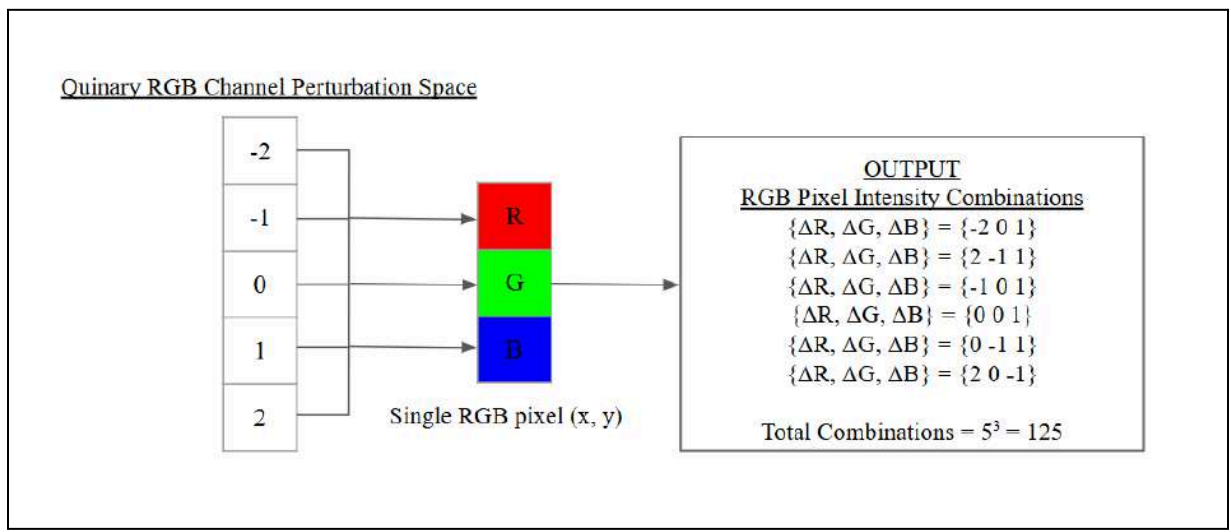

Fig 1. Quinary RGB Channel Perturbation space in a single RGB pixel

Once the quinary RGB perturbation space is defined, textual information is embedded by assigning each character to a unique perturbation triplet. The character set covers uppercase letters (A-Z), lowercase letters (a-z), numerals (0-9), and commonly used special characters, allowing direct representation of text without additional encoding or compression. Each symbol corresponds to exactly one valid quinary combination, ensuring deterministic embedding and unambiguous decoding. The complete character-to-perturbation mapping used in this work is summarized in appendix A and the mapping is realized by applying the corresponding perturbations independently to the R, G, and B channels of the selected pixel, as illustrated in the figure 2.

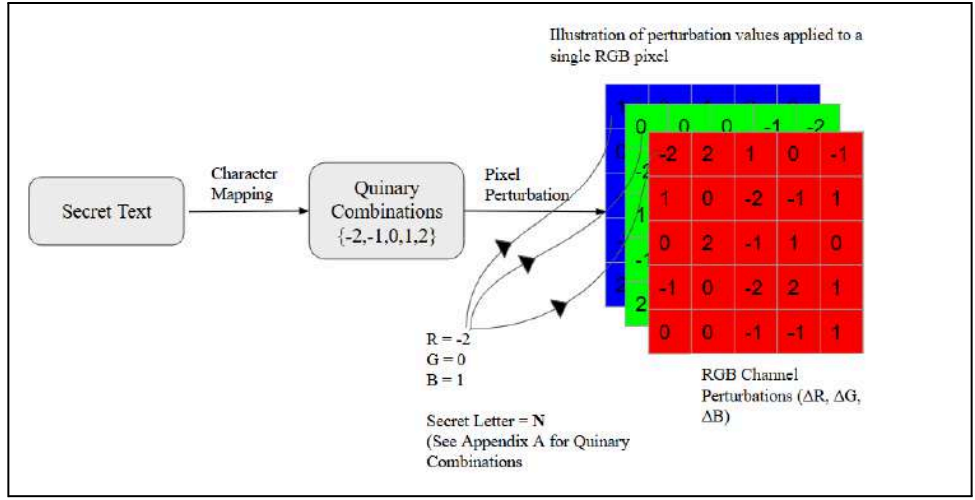

**Fig 2. Pixel Perturbation in RGB Image using Quinary Combinations.**

## 2.1. MATHEMATICAL FORMULATION:
Let the cover image be denoted as
$$I \in Z^{H \times W \times 3}$$
where each pixel at spatial location (x,y) is represented as
$$P(x,y) = (R(x,y), G(x,y), B(x,y))$$
with $R,G,B \in [0,255]$.
and the quinary perturbation set as
$$Q = \{-2,-1,0,+1,+2\}.$$
Each textual symbol 's' from the predefined character set 'S' is mapped via a deterministic mapping function
$$f : S \rightarrow Q^3$$
such that
$$f(s) = (\Delta R_s, \Delta G_s, \Delta B_s),$$
where $\Delta R_s, \Delta G_s, \Delta B_s \in Q$.
This mapping yields up to $|Q|^3 = 125$ distinct symbol representations.
### 2.1.1. Encoding:
Given a selected pixel P(x,y) and a symbol 's', the Stego pixel P′(x,y) is computed as
$$P'(x,y) = P(x,y) + f(s),$$
or explicitly,
$$R'(x,y) = R(x,y) + \Delta R_s,$$

$$G'(x,y) = G(x,y) + \Delta G_s,$$
$$B'(x,y) = B(x,y) + \Delta B_s.$$

To ensure valid pixel intensities, the embedding is subject to the constraint
$$0 \leq R', G', B' \leq 255.$$

If a perturbation violates this constraint, a sign inverted perturbation or pixel skipping strategy is applied while preserving decoding consistency.

### 2.1.2. Decoding:

Decoding assumes availability of the original cover image. For a Stego pixel $P'(x,y)$, the perturbation vector is recovered as
$$\Delta P(x,y) = P'(x,y) - P(x,y).$$

The original symbol '$\hat{s}$' is then obtained by inverse mapping:
$$\hat{s} = f^{-1}(\Delta P(x,y)).$$

Since the mapping 'f' is injective over the selected symbol set, decoding is deterministic and lossless, provided the perturbation constraints are satisfied.

## 3. EXPERIMENTAL SETUP:

The experimental setup evaluates the proposed quinary perturbation framework by embedding textual content of varying lengths, categorized as single-paragraph and multi-paragraph payloads, into images of different spatial resolutions. Textual payloads are injected into different image resolutions such as 512×512, 1280×780, 1920×1080, and 3840×2160 to study the behavior of the method under increasing payload density and image size. For each case, the encoded images are generated using the same perturbation and mapping strategy, and the decoded text is verified for lossless recovery. The quality and imperceptibility of the encoded images are analyzed using quantitative metrics including MSE, MAE, SNR, PSNR, and SSIM, while Heatmap analysis and Histogram Comparison are used to examine pixel level perturbation patterns and intensity distributions. Figure 3 shows the experimental datasets used in this work.

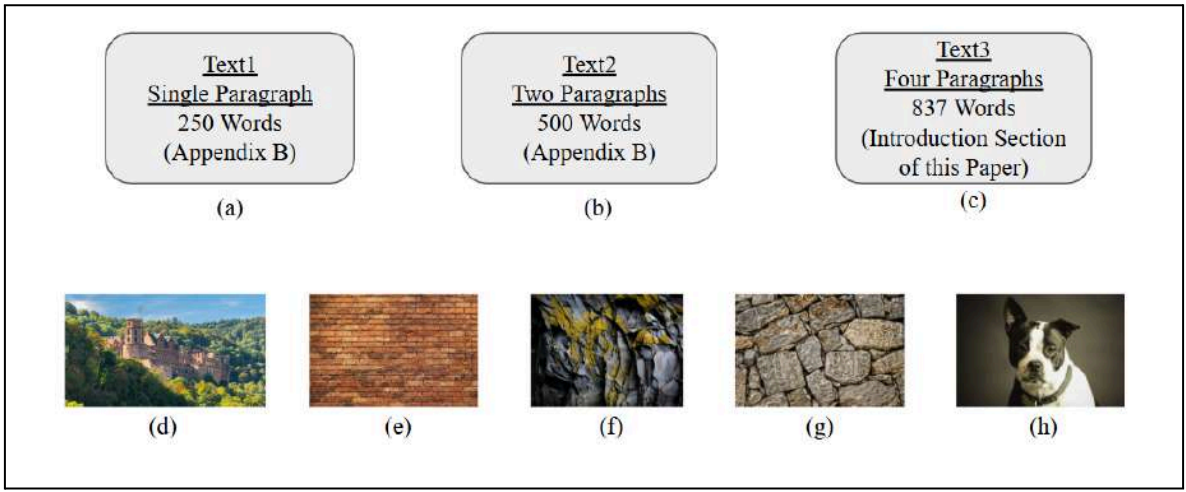

**Fig 3. Datasets used in this work, (a),(b),(c): Textual Data, (d),(e),(f),(g),(h): Image Datasets (See Appendix B)**

## 4. EXPERIMENTS & RESULTS:

To evaluate the effectiveness, scalability, and robustness of the proposed quinary pixel combination based text embedding framework, a series of experiments were conducted across multiple image resolutions using a single representative cover image. The same textual payload comprising the complete introduction section of this paper (Text3) as stated above, was embedded into images of four different spatial resolutions such as 512×512, 1280×720, 1920×1080, and 3840×2160. This experimental design allows direct analysis of the method's behavior under increasing image capacity while maintaining a constant payload size. For each resolution, the Stego image was generated using identical quinary perturbation mappings, and the embedded text was decoded deterministically using the original cover image as reference. Quantitative evaluation was performed using standard image quality metrics, including MAE, MSE, SNR, PSNR, and SSIM, along with character and word level recovery accuracy (CER/WER). The results, summarized in Table 1, demonstrate consistently high visual fidelity across all resolutions, with increasing PSNR and SNR values as image size grows, while maintaining near perfect structural similarity. In parallel, Table 2 reports payload utilization statistics, highlighting the extremely low fraction of modified pixels required to embed the full textual payload, particularly at higher resolutions. Additionally, histogram comparison and heatmap analyses show minimal intensity distribution shifts and spatially sparse perturbations, confirming that the proposed method introduces no visually discernible artifacts across all tested configurations. The following figures 4-7, shows the original & decoded images, along with payload utilization metrics, histogram comparisons and heatmaps.

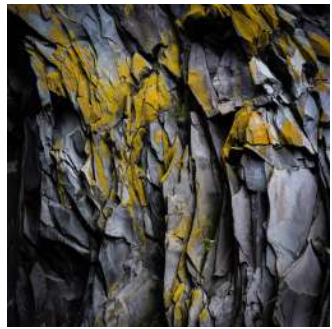 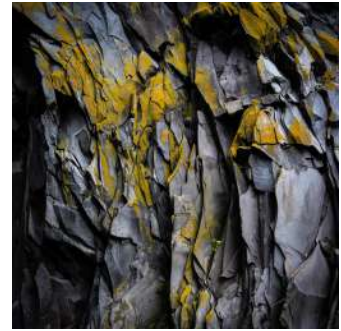

(a) (b)

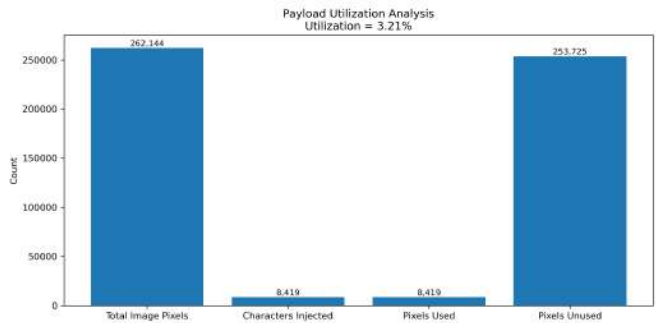

(c)

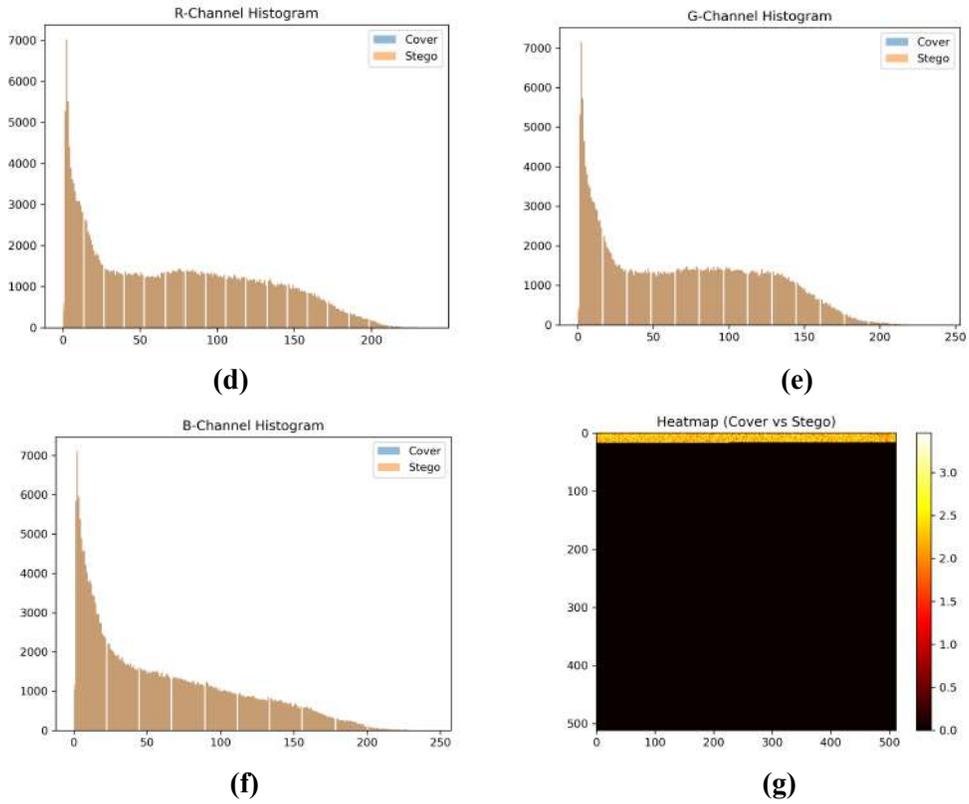

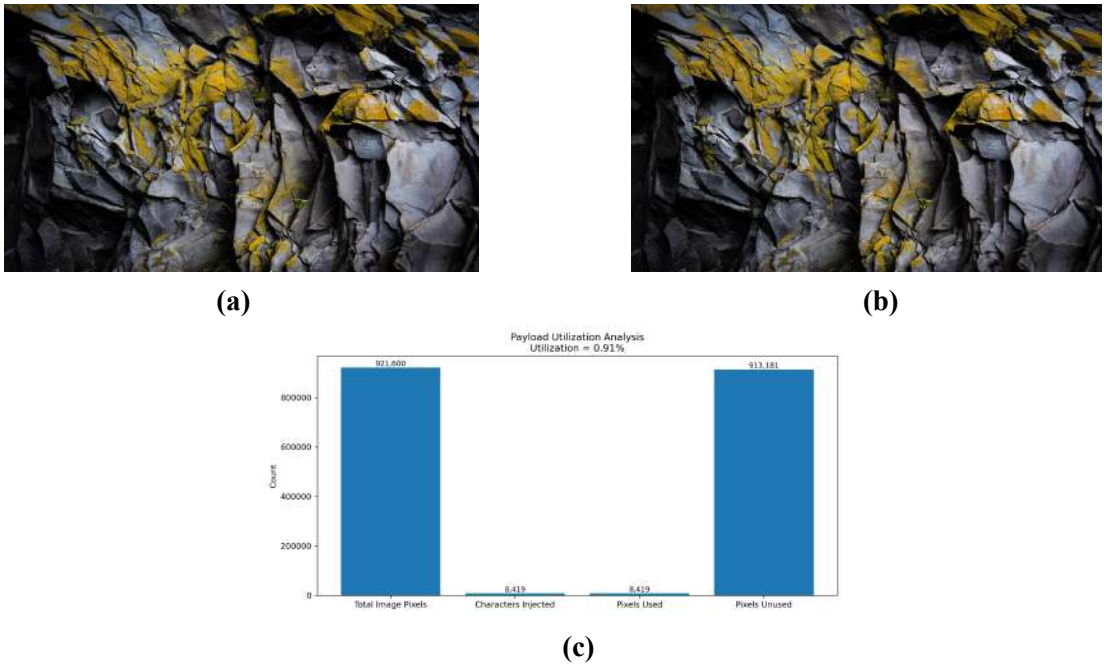

**Fig 4.** Shows (a) Original Image & (b) Encoded (Stego) Image of resolution 512×512, (c) Payload Utilization Plot, (d),(e)&(f) are Histograms for cover and Stego images, (g) Heatmap between Cover & Stego Images.

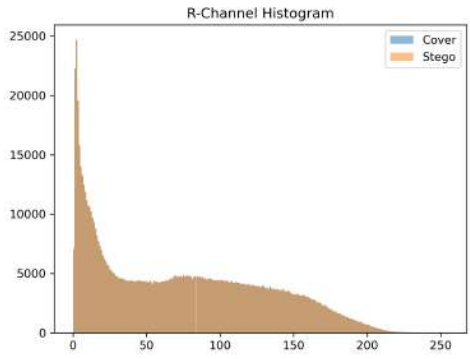

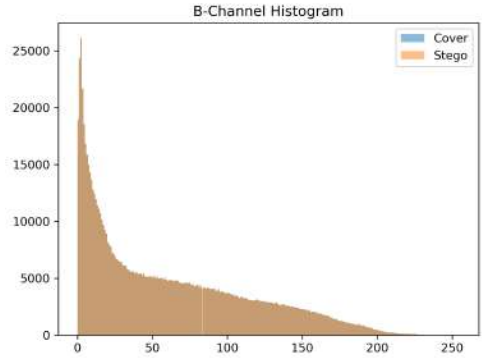

(d)

(e)

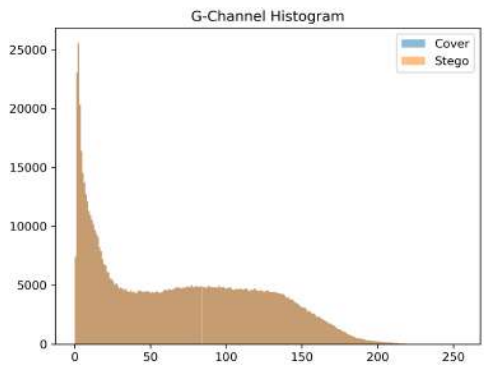

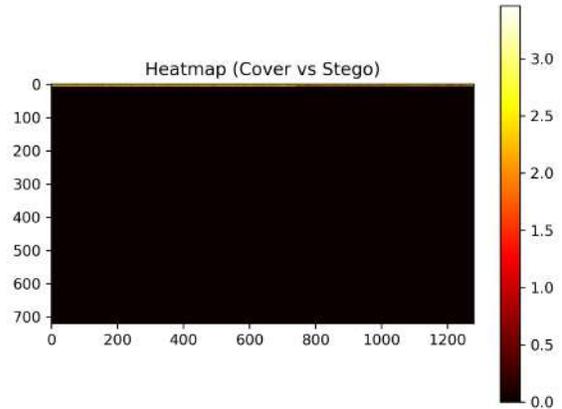

(f)

(g)

**Fig 5. Shows (a) Original Image & (b) Encoded (Stego) Image of resolution 1280×720, (c) Payload Utilization Plot, (d),(e)&(f) are Histograms for cover and Stego images, (g) Heatmap between Cover & Stego Images.**

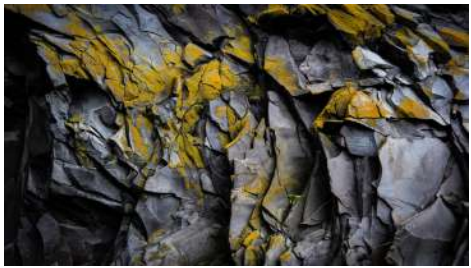

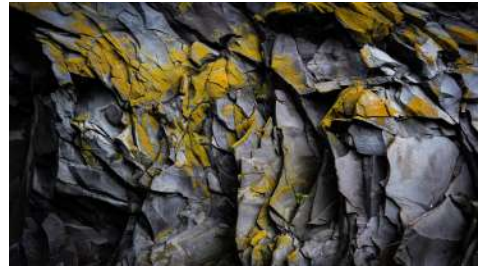

(a)

(b)

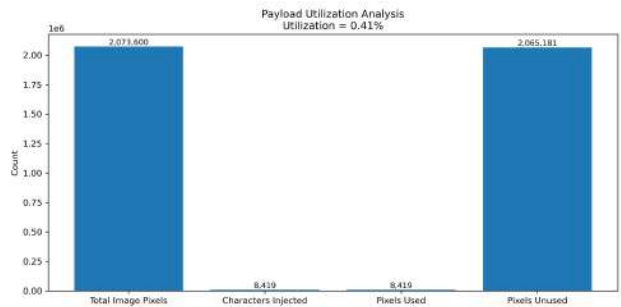

(c)

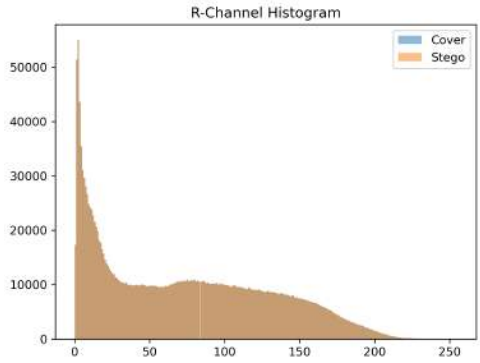
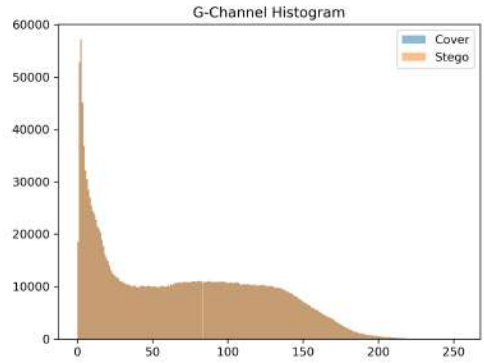

(d) (e)

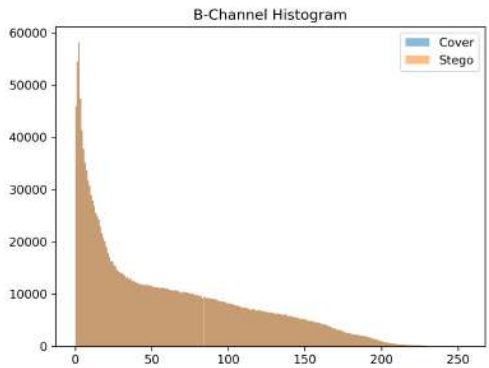
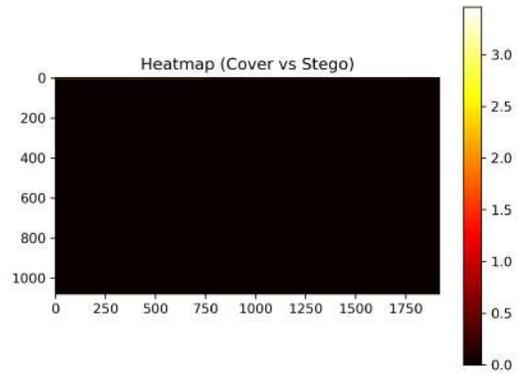

(f) (g)

**Fig 6. Shows (a) Original Image & (b) Encoded (Stego) Image of resolution 1920×1080, (c) Payload Utilization Plot, (d),(e)&(f) are Histograms for cover and Stego images, (g) Heatmap between Cover & Stego Images.**

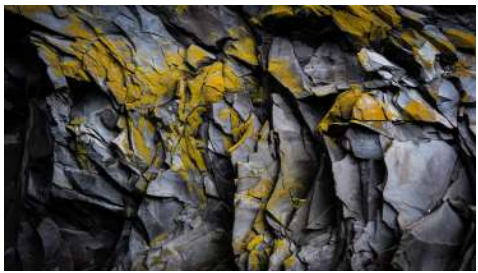
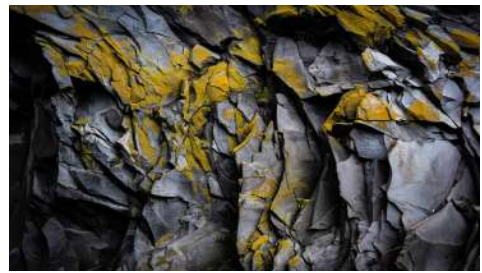

(a) (b)

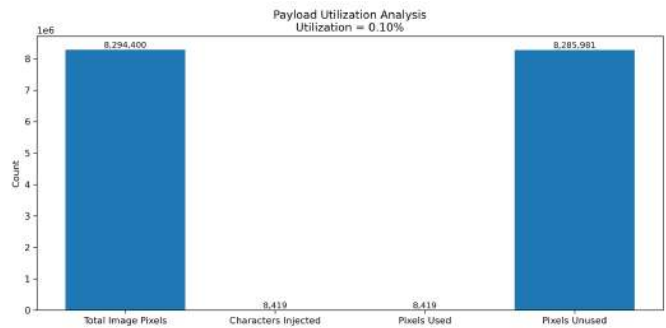

(c)

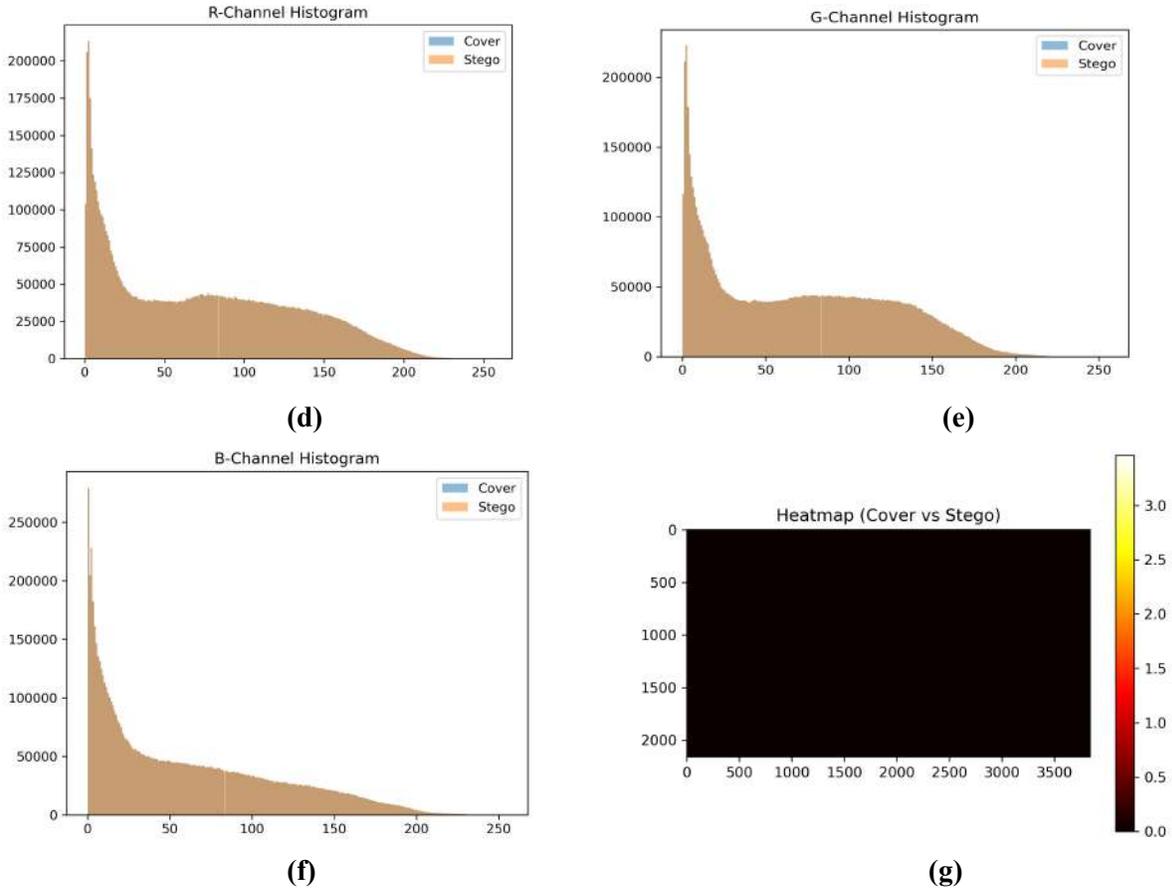

(d) (e)

(f) (g)

Fig 7. Shows (a) Original Image & (b) Encoded (Stego) Image of resolution 3840×2160, (c) Payload Utilization Plot, (d),(e)&(f) are Histograms for cover and Stego images, (g) Heatmap between Cover & Stego Images.

| IMAGE RESOLUTION | MAE | MSE | SNR | PSNR | SSIM | CER / WER |
|---|---|---|---|---|---|---|
| 512×512 | 0.039 | 0.0584 | 51.042 | 60.465 | 0.999 | 0.0003/0.002 |
| 1280×720 | 0.0109 | 0.0163 | 56.634 | 65.989 | 0.999 | 0.0003/0.002 |
| 1920×1080 | 0.0048 | 0.0072 | 60.15 | 69.513 | 0.999 | 0.0003/0.002 |
| 3840×2160 | 0.00179 | 0.0012 | 66.245 | 75.582 | 0.999 | 0.0003/0.002 |

Table 1. Shows MAE, MSE, SNR, PSNR, SSIM & CER / WER of different resolutions of the same Cover Image before (original) and after encoding (Stego).

| IMAGE RESOLUTION | Total Pixels | Payload Count | Pixels Used | Pixels Unused | Payload Utilization Percentage |
|---|---|---|---|---|---|
| 512×512 | 262144 | 8419 | 8419 | 253725 | 3.211 |
| 1280×720 | 921600 | 8419 | 8419 | 913181 | 0.9135 |
| 1920×1080 | 2073600 | 8419 | 8419 | 2065181 | 0.406 |
| 3840×2160 | 8294400 | 8419 | 8419 | 8285981 | 0.1015 |

**Table 2. Shows Total pixels, Payload Count, Pixels Used, Pixels, Unused and Pixel Utilization Percentage of different resolutions of the same Cover Image before (original) and Payload (Secret Data).**

The experimental results clearly demonstrate that the proposed quinary pixel combination–based embedding framework achieves high visual imperceptibility and reliable text recovery across all tested image resolutions. As shown in Table 1, image distortion metrics consistently remain within negligible ranges, with PSNR and SNR values increasing as image resolution grows due to reduced payload density, while SSIM values remain close to unity in all cases. This indicates that the controlled quinary perturbations preserve both pixel-level fidelity and structural characteristics of the cover images. Table 2 further highlights the symbol-level efficiency of the method, revealing that the complete textual payload is embedded using a very small fraction of available pixels, particularly for higher resolutions where payload utilization drops below 1%. Histogram comparisons and heatmap visualizations corroborate these findings by exhibiting minimal distributional shifts and spatially sparse perturbation patterns, confirming the absence of visually discernible artifacts. Collectively, these results validate the scalability, determinism, and practical effectiveness of the proposed approach for text embedding in images.

To further support the generality of the framework, additional experiments using different textual payloads, including shorter and medium length paragraphs with varied formatting, are provided in the Appendix B. These supplementary results reinforce the observed trends in image quality preservation and text recovery accuracy, and demonstrate consistent behavior of the proposed method across diverse payload characteristics.

In comparison with conventional spatial-domain text embedding techniques such as LSB, MSB, and Pixel Value Differencing (PVD), the proposed method demonstrates clear advantages in terms of spatial efficiency. Traditional bit-based approaches typically require modifying multiple pixels to represent a single textual character, a minimum of 8 bits across several pixels or channels to reconstruct one symbol as shown in the Table 3. As a result, the payload is spatially dispersed, increasing the number of modified pixels and the overall footprint of embedding. In contrast, the proposed quinary pixel combination framework encodes a complete textual symbol directly within a single RGB pixel by jointly modulating its three color channels. This symbol-level encoding significantly reduces the number of pixels required per character, leading to substantially lower pixel utilization rates, as reflected in Table 2. Consequently, the method achieves higher spatial efficiency and reduced perturbation density compared to traditional LSB, MSB, and PVD based schemes, while maintaining high visual fidelity and deterministic decoding.

Although the proposed framework is demonstrated on RGB images, the symbol-level quinary encoding can be directly extended to grayscale images by distributing the three perturbation components across three grayscale pixels. In this case, a complete textual character is embedded using only three grayscale pixels, which remains significantly more spatially efficient than conventional LSB or MSB based grayscale embedding schemes that typically require eight pixels per character.

| METHOD | PIXELS PER CHARACTER |
|---|---|
| LSB (Grayscale) | 8 |
| LSB (RGB, 1-bit/channel) | 3 |
| MSB (Grayscale/RGB) | 8/3 |
| Proposed Method (Grayscale / RGB) | 3/1 |

Table 3. Comparison of Pixel Usage per Character in Spatial Domain Methods

## 5. LIMITATIONS & FUTURE SCOPE:

The proposed quinary pixel combination based text embedding framework is designed for deterministic, high efficiency spatial-domain steganography under controlled conditions. As decoding relies on direct pixel-wise comparison with the original cover image, the method assumes availability of the unmodified cover image at the receiver side. Consequently, the current implementation is not inherently robust to lossy image transformations such as aggressive compression, resizing, or post-processing operations that alter pixel intensities beyond the bounded quinary perturbation range. In addition, the present symbol mapping is defined over a fixed character set, which necessitates basic text normalization when handling typographic Unicode characters or non-standard formatting commonly encountered in various texts. These constraints are inherent to deterministic spatial-domain symbol-level embedding and are not unique to the proposed approach.

The proposed framework opens multiple directions for future research and extension. First, robustness against common image transformations can be enhanced by integrating adaptive pixel selection strategies or lightweight error-correcting mechanisms while preserving symbol-level efficiency. Second, the quinary perturbation space can be extended to support richer character sets, including full Unicode mappings or domain-specific symbol vocabularies, enabling broader applicability. Third, adaptive embedding strategies that exploit local image characteristics such as texture or edge density may further reduce detectability while maintaining deterministic decoding. Additionally, the proposed symbol-level encoding paradigm can be generalized to other data modalities, such as embedding structured metadata, sensor logs, audio and video data, or command tokens within images for secure communication in resource constrained or cyber-physical systems. These extensions position the proposed approach as a foundational framework for efficient, interpretable, and scalable spatial-domain steganography.

## 6. CONCLUSION:

This paper presented a deterministic and spatially efficient image steganography framework based on quinary pixel intensity combinations in RGB space for embedding textual information. By constraining each color channel to five controlled perturbation levels, the proposed method enables direct symbol-level encoding, allowing a complete textual character to be represented within a single RGB pixel. This design departs from traditional bit-centric spatial-domain techniques and achieves significantly higher embedding efficiency while maintaining strong visual imperceptibility.

Comprehensive experiments conducted across multiple image resolutions, ranging from 512×512 to 3840×2160, demonstrate that the proposed approach preserves image quality with negligible distortion, as reflected by consistently high PSNR, SNR, and SSIM values. Payload utilization analysis further highlights the method's spatial efficiency, with only a small fraction of available pixels required to embed a complete academic-length textual payload, particularly at higher resolutions. Deterministic decoding was achieved in all cases, with character and word level accuracy remaining consistently high, and minor discrepancies attributable to text normalization rather than embedding failure.

By combining simplicity, interpretability, and symbol-level efficiency, the proposed quinary pixel combination framework provides a lightweight and practical alternative to conventional LSB, MSB, PVD, and learning-based steganographic methods. The results establish the method as a scalable and effective solution for text embedding in images, suitable for secure communication and information hiding scenarios where computational efficiency, deterministic recovery, and minimal visual distortion are essential.

**CODE AVAILABILITY:**
The reference implementation of this work, along with scripts for reproducing all experiments reported in this paper, is publicly available at:
https://github.com/Udaykandhala/Quinary-Pixel-Combination-Steganography.

**ACKNOWLEDGEMENT:**
This research was conducted independently by the author. No external funding, institutional grants, or departmental resources were used in the development of the framework, methodology, or experimental results etc. For polishing the original draft for better language, contextual understanding and for code assistance, the author used generative AI language tools such as ChatGPT, Gemini in this work.

**CONFLICTS OF INTERESTS:**
The author declares that there are no conflicts of interests with any persons or organizations that could influence this work.

**AUTHOR CONTRIBUTIONS:**
A V Uday Kiran Kandala is the sole contributor to the conceptualization, methodology, software development, paper drafting, editing and formal analysis of this work.

**APPENDIX A:**
The ground truth of quinary pixel intensity combinations for character mapping used in this work. A total of 98 different R,G,B pixel intensity combinations are used in this work, and 27 combinations are shelved for future scope or for additional character mappings in future work.

| Character | Delta_R | Delta_G | Delta_B | Character | Delta_R | Delta_G | Delta_B |
|---|---|---|---|---|---|---|---|
| A | -2 | -2 | -2 | a | -1 | -2 | -1 |
| B | -2 | -2 | -1 | b | -1 | -2 | 0 |
| C | -2 | -2 | 0 | c | -1 | -2 | 1 |
| D | -2 | -2 | 1 | d | -1 | -2 | 2 |
| E | -2 | -2 | 2 | e | -1 | -1 | -2 |
| F | -2 | -1 | -2 | f | -1 | -1 | -1 |
| G | -2 | -1 | -1 | g | -1 | -1 | 0 |
| H | -2 | -1 | 0 | h | -1 | -1 | 1 |
| I | -2 | -1 | 1 | i | -1 | -1 | 2 |
| J | -2 | -1 | 2 | j | -1 | 0 | -2 |
| K | -2 | 0 | -2 | k | -1 | 0 | -1 |
| L | -2 | 0 | -1 | l | -1 | 0 | 0 |
| M | -2 | 0 | 0 | m | -1 | 0 | 1 |
| N | -2 | 0 | 1 | n | -1 | 0 | 2 |
| O | -2 | 0 | 2 | o | -1 | 1 | -2 |
| P | -2 | 1 | -2 | p | -1 | 1 | -1 |
| Q | -2 | 1 | -1 | q | -1 | 1 | 0 |
| R | -2 | 1 | 0 | r | -1 | 1 | 1 |
| S | -2 | 1 | 1 | s | -1 | 1 | 2 |
| T | -2 | 1 | 2 | t | -1 | 2 | -2 |
| U | -2 | 2 | -2 | u | -1 | 2 | -1 |
| V | -2 | 2 | -1 | v | -1 | 2 | 0 |
| W | -2 | 2 | 0 | w | -1 | 2 | 1 |
| X | -2 | 2 | 1 | x | -1 | 2 | 2 |
| Y | -2 | 2 | 2 | y | 0 | -2 | -2 |
| Z | -1 | -2 | -2 | z | 0 | -2 | -1 |

| Character | Delta_R | Delta_G | Delta_B | Character | Delta_R | Delta_G | Delta_B |
|---|---|---|---|---|---|---|---|
| **0** | 0 | -2 | 0 | **{** | 1 | -2 | 1 |
| **1** | 0 | -2 | 1 | **}** | 1 | -2 | 2 |
| **2** | 0 | -2 | 2 | **\|** | 1 | -1 | -2 |
| **3** | 0 | -1 | -2 | **[** | 1 | -1 | -1 |
| **4** | 0 | -1 | -1 | **]** | 1 | -1 | 0 |
| **5** | 0 | -1 | 0 | **\\** | 1 | -1 | 1 |
| **6** | 0 | -1 | 1 | **:** | 1 | -1 | 2 |
| **7** | 0 | -1 | 2 | **'** | 1 | 0 | -2 |
| **8** | 0 | 0 | -2 | **;** | 1 | 0 | -1 |
| **9** | 0 | 0 | -1 | **"** | 1 | 0 | 0 |
| **`** | 0 | 0 | 0 | **<** | 1 | 0 | 1 |
| **~** | 0 | 0 | 1 | **>** | 1 | 0 | 2 |
| **!** | 0 | 0 | 2 | **?** | 1 | 1 | -2 |
| **@** | 0 | 1 | -2 | **,** | 1 | 1 | -1 |
| **#** | 0 | 1 | -1 | **.** | 1 | 1 | 0 |
| **$** | 0 | 1 | 0 | **/** | 1 | 1 | 1 |
| **%** | 0 | 1 | 1 | **Whitespace** | 1 | 1 | 2 |
| **^** | 0 | 1 | 2 | **\n** | 1 | 2 | -2 |
| **&** | 0 | 2 | -2 | **\n\n** | 1 | 2 | -1 |
| **\*** | 0 | 2 | -1 | **NUL** | 1 | 2 | 0 |
| **(** | 0 | 2 | 0 | | | | |
| **)** | 0 | 2 | 1 | | | | |
| **_** | 0 | 2 | 2 | | | | |
| **+** | 1 | -2 | -2 | | | | |
| **-** | 1 | -2 | -1 | | | | |
| **=** | 1 | -2 | 0 | | | | |

**APPENDIX B (ADDITIONAL EXPERIMENTS):**
The appendix shows the additional experiments of single (TEXT1) and multiple para (TEXT2 & 3) injections into 512×512, 1280×780, 1920×1080, and 3840×2160 resolution images.

**512×512 (TEXT2):**

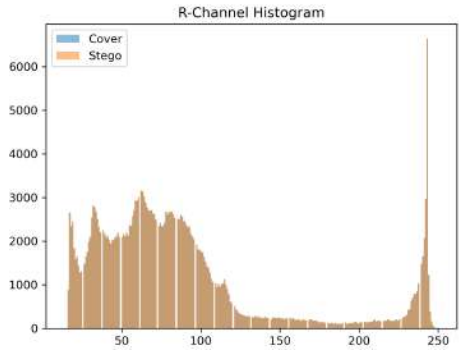
(a)

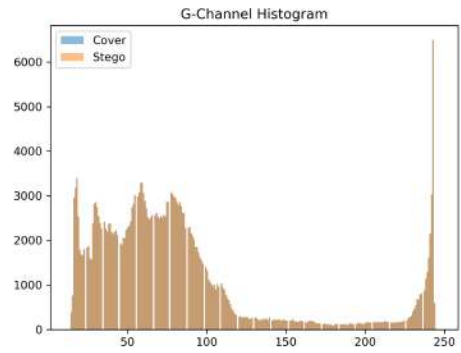
(b)

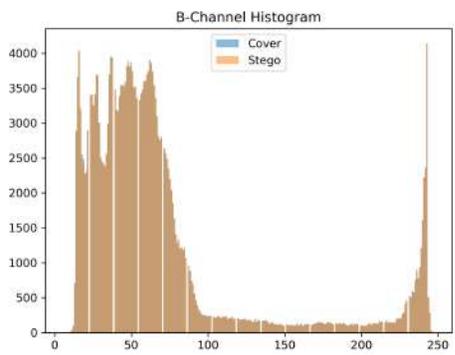
(c)

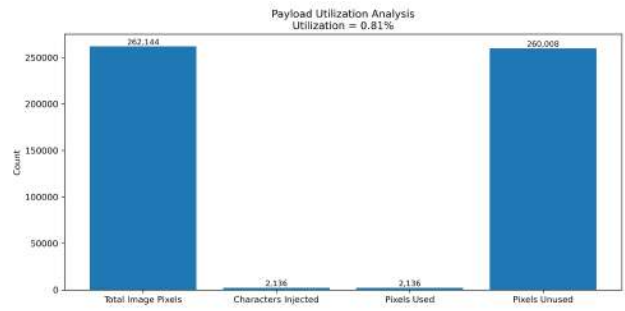
(d)

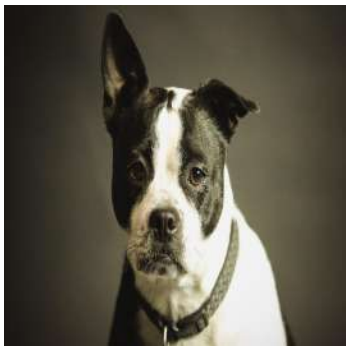
(e)

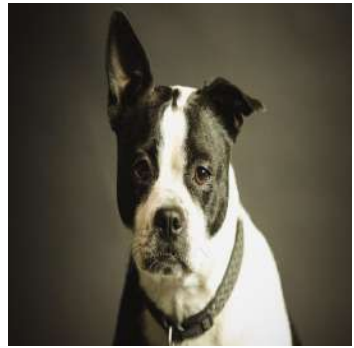
(f)

**Fig 8. (a),(b),(c), Histogram Comparison of R,G,B, channels for original and encoded images, (d), Payload Utilization Analysis plot, (e)&(f) Original and Encoded Images.**

**1280×780 (TEXT2):**

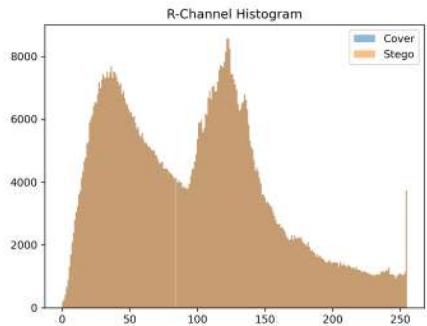
(a)

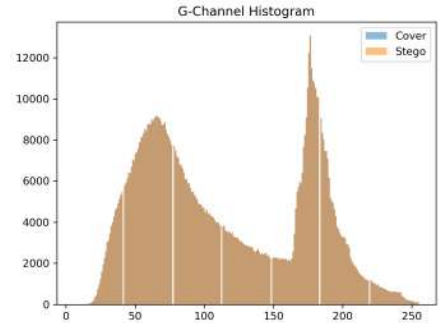
(b)

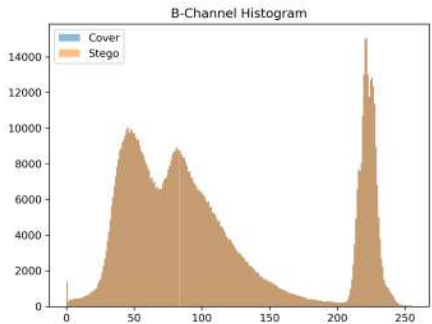
(c)

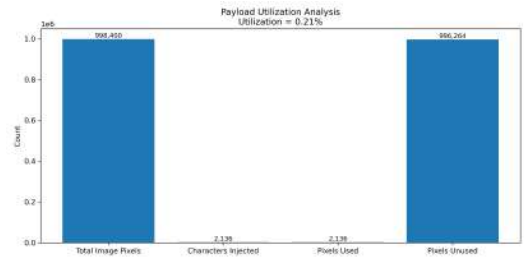
(d)

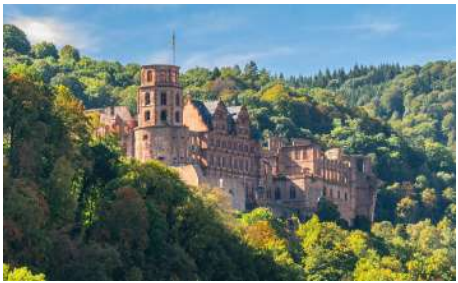
(e)

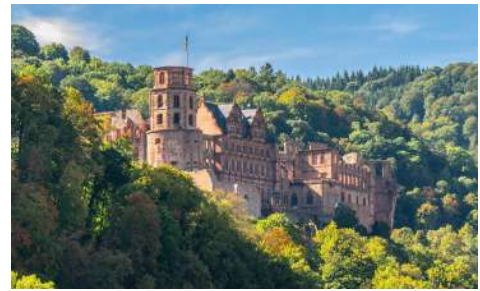
(f)

**Fig 9. (a),(b),(c), Histogram Comparison of R,G,B, channels for original and encoded images, (d), Payload Utilization Analysis plot, (e)&(f) Original and Encoded Images.**

**1920×1080 (TEXT2):**

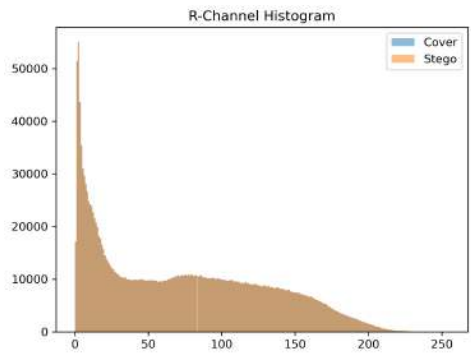
(a)

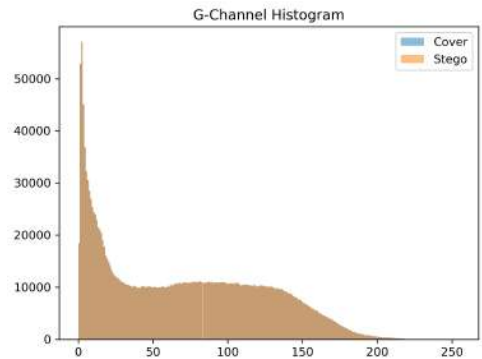
(b)

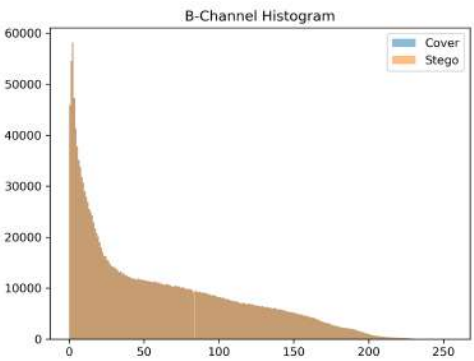
(c)

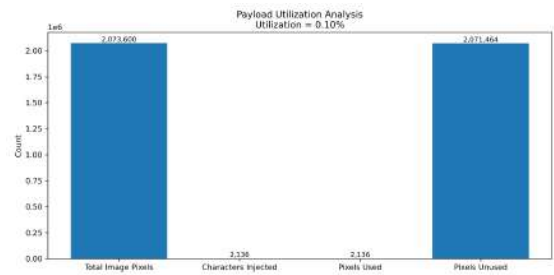
(d)

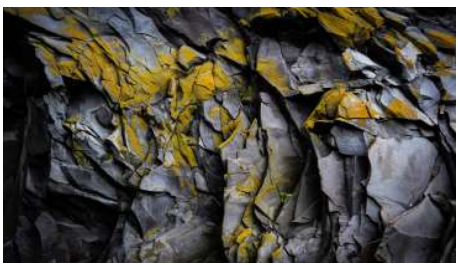
(e)

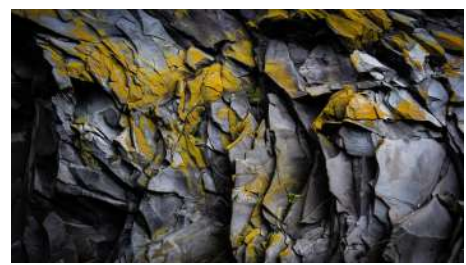
(f)

Fig 10. (a),(b),(c), Histogram Comparison of R,G,B, channels for original and encoded images, (d), Payload Utilization Analysis plot, (e)&(f) Original and Encoded Images.

**3840×2160 (TEXT1):**

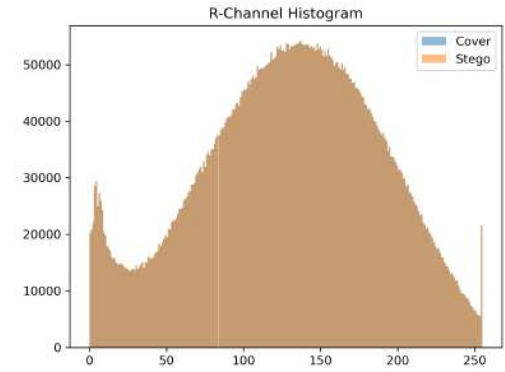

(a)

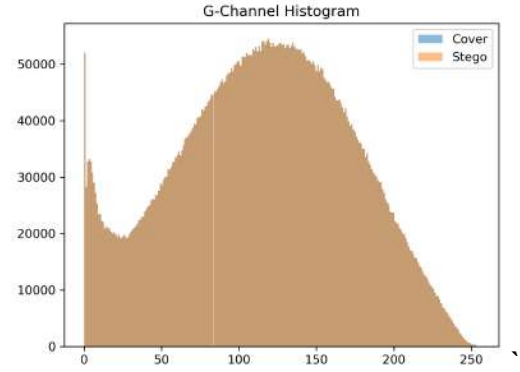

(b)

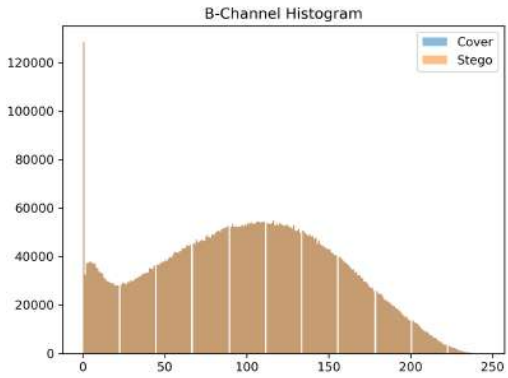

(c)

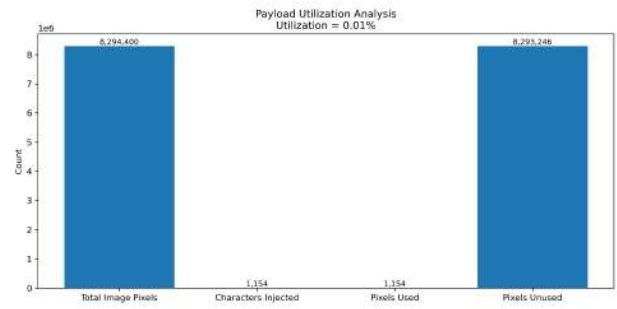

(d)

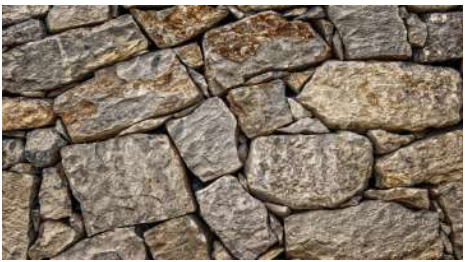

(e)

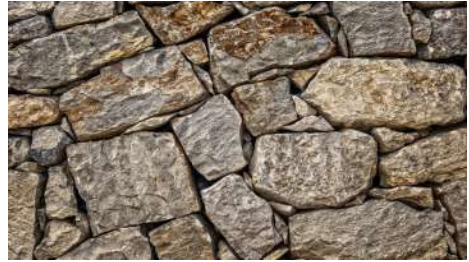

(f)

**Fig 11. (a),(b),(c), Histogram Comparison of R,G,B, channels for original and encoded images, (d), Payload Utilization Analysis plot, (e)&(f) Original and Encoded Images.**

**APPENDIX C:**

The Text datasets used in this work:

**TEXT1 (Generated by ChatGPT-250WORDS):**

On a quiet Sunday morning, Arun woke up at 7:30 AM and checked his phone. The message count showed 12 unread notifications, but none of them felt urgent. Outside, the sky was clear, and the temperature was around 24.5 degrees. It felt like a good day to slow down.

He made a cup of coffee, added 2 spoons of sugar, and sat near the window. Cars passed by slowly, people walked their dogs, and children laughed near the park. Life felt ordinary, and that was comforting. Arun opened his notebook and wrote a small reminder: "Finish tasks, but don't rush."

By noon, he visited a local bookstore. The price tag on one book read $19.99, another showed $25.00, and a third was on sale for $10. He smiled and picked the cheapest one. The cashier asked, "Cash or card?" Arun replied, "Card, please."

In the evening, he called his sister and spoke for 15 minutes. They talked about work, travel plans, and simple memories from school. At 9:45 PM, Arun closed his laptop, set an alarm for 6:00 AM, and turned off the lights.

Before sleeping, he thought about the day. Nothing dramatic happened, no big success or failure. Still, the balance of small moments felt meaningful.

**TEXT2 (Generated by ChatGPT-500WORDS):**

On a quiet weekday morning, Rohan woke up at 6:45 AM and looked out through his bedroom window. The street below was calm, with only a few people walking to work and a couple of shops opening their shutters. He checked his phone and noticed 18 new messages, most of them reminders, updates, and simple greetings. Nothing felt urgent, and that gave him a sense of ease.

After brushing his teeth and making breakfast, Rohan prepared a cup of tea. He added milk, sugar, and stirred it slowly while listening to the morning news. The temperature outside was around 26 degrees, and the weather forecast predicted a clear day. He made a short list of tasks in his notebook: reply to emails, review documents, attend one meeting, and go for an evening walk.

By mid-morning, Rohan was working on his laptop at the dining table. The clock showed 10:30 AM, and sunlight filled the room. His internet connection briefly dropped for about 2 minutes, but it returned without any issue. He continued reading a report that contained dates, numbers, and references. Some sections were marked important, while others could wait until later.

At lunchtime, Rohan ordered food from a nearby restaurant. The total bill came to $12.75, including taxes. While waiting, he read a few pages from a book he bought last week for $18. The story was simple, but it helped him relax and step away from work pressure. When the food arrived, he ate slowly and avoided checking his phone.

In the afternoon, a colleague called him to discuss a small problem. The call lasted about 14 minutes, and they agreed on a solution without much debate. Rohan updated his notes, saved his files, and backed up important data before shutting down his system at 5:45 PM.

As evening approached, Rohan went outside for a walk. Cars passed by, streetlights turned on, and the sky slowly changed color. At 8:00 PM, he returned home, cooked a light dinner, and watched a short documentary. Before sleeping, he set an alarm for 6:00 AM, checked that everything was in order, and turned off the lights. The day ended quietly, without stress or excitement, but with a steady sense of completion.

**TEXT3:**

The complete introduction para of this work.

## IMAGE RESOURCES:

The images used in this work are taken from open source freely available images from Pixabay, the related links of those images are as follows for credits.

Image by <a href="https://pixabay.com/users/tama66-1032521/?utm_source=link-attribution&utm_medium=referral&utm_campaign=image&utm_content=3630911">Peter H</a> from <a href="https://pixabay.com//?utm_source=link-attribution&utm_medium=referral&utm_campaign=image&utm_content=3630911">Pixabay</a>

Image by <a href="https://pixabay.com/users/analogicus-8164369/?utm_source=link-attribution&utm_medium=referral&utm_campaign=image&utm_content=6961638">Tom</a> from <a href="https://pixabay.com//?utm_source=link-attribution&utm_medium=referral&utm_campaign=image&utm_content=6961638">Pixabay</a>

Image by <a href="https://pixabay.com/users/kie-ker-2367988/?utm_source=link-attribution&utm_medium=referral&utm_campaign=image&utm_content=1346727">esiuL</a> from <a href="https://pixabay.com//?utm_source=link-attribution&utm_medium=referral&utm_campaign=image&utm_content=1346727">Pixabay</a>

Image by <a href="https://pixabay.com/users/gruendercoach-13177285/?utm_source=link-attribution&utm_medium=referral&utm_campaign=image&utm_content=8540772">Siegfried Poepperl</a> from <a href="https://pixabay.com//?utm_source=link-attribution&utm_medium=referral&utm_campaign=image&utm_content=8540772">Pixabay</a>

Image by <a href="https://pixabay.com/users/pexels-2286921/?utm_source=link-attribution&utm_medium=referral&utm_campaign=image&utm_content=2178696">Pexels</a> from <a href="https://pixabay.com//?utm_source=link-attribution&utm_medium=referral&utm_campaign=image&utm_content=2178696">Pixabay</a>

**-END OF THE PAPER-**